\documentclass[letterpaper, 10 pt, conference]{ieeeconf}  

\IEEEoverridecommandlockouts                              

\usepackage{graphics} 
\usepackage{epsfig} 
\usepackage{amsmath} 
\usepackage{amssymb}  
\usepackage{algorithm}
\usepackage{verbatim}
\usepackage[noend]{algpseudocode}
\usepackage{subfig}
\usepackage{multirow}
\usepackage{amsthm}

\let\eqref\undefined
\newcommand{\figref}[1]{Fig.~\ref{fig:#1}}

\newcommand{\tabref}[1]{Table~\ref{tab:#1}}
\newcommand{\eqref}[1]{Eqn.~\ref{eq:#1}}

\newcommand{\eg}[1]{\textit{e.g.,}}
\newcommand{\etc}[1]{\textit{etc.}}
\newcommand{\vs}[1]{\textit{vs.}}

\theoremstyle{definition}

\newtheorem{theorem}{Theorem}

\newtheorem{lemma}{Lemma}

\title{\LARGE \bf Localization under Topological Uncertainty for Lane Identification of Autonomous Vehicles}

\author{Samer B. Nashed$^{1, 2}$, David M. Ilstrup$^{2}$, and Joydeep Biswas$^{1}$ 
\thanks{$^{1}$Samer B. Nashed and Joydeep Biswas are with the College of Information and Computer Sciences,
        University of Massachusetts, Amherst, MA 01003, USA.
      Email: {\tt\small \{snashed, joydeepb\}@cs.umass.edu}
   $^{2}$ David M. Ilstrup is at Nissan Research Center,
       1215 Bordeaux Drive, Sunnyvale, CA 94089
     Email: {\tt\small \{david.ilstrup\}@nissan-usa.com}}}%

\begin{document}

\bibliographystyle{abbrv}

\maketitle
\thispagestyle{empty}
\pagestyle{empty}

\begin{abstract}
Autonomous vehicles (AVs) require accurate metric and topological location estimates for safe, effective navigation and decision-making. Although many high-definition (HD) roadmaps exist, they are not always accurate since public roads are dynamic, shaped unpredictably by both human activity and nature. Thus, AVs must be able to handle situations in which the topology specified by the map does not agree with reality. We present the Variable Structure Multiple Hidden Markov Model (VSM-HMM) as a framework for localizing in the presence of topological uncertainty, and demonstrate its effectiveness on an AV where lane membership is modeled as a topological localization process. VSM-HMMs use a dynamic set of HMMs to simultaneously reason about location within a set of most likely current topologies and therefore may also be applied to topological structure estimation as well as AV lane estimation. In addition, we present an extension to the Earth Mover's Distance which allows uncertainty to be taken into account when computing the distance between belief distributions on simplices of arbitrary relative sizes.
\end{abstract}

\section{Introduction}

\vspace{-1mm}

Localization is an essential capability for autonomous mobile robots, including autonomous vehicles (AVs). Most localization algorithms use metric maps as aids in the localization process, which represent features in continuous coordinates. Topological maps represent space as discrete components (vertices) and their logical-spatial relationship (edges) where vertices and edges are taken in the graph theoretic sense. The motivating example in this paper is an AV which, in addition to requiring a metric location estimate, also requires a topological location estimate at the lane level. For example, the AV may need to know not just that it is on Pleasant Street, but whether or not it is in the left turn only lane. Moreover, events such as construction, traffic accidents, natural disasters, and native map errors may result in discrepancies between the topology suggested by the map and reality. The localization algorithm on the AV must be able to reason about this possibility. 

Reasoning \emph{globally} about all possible topologies is computationally intractable, since the number of unique topologies scales exponentially with the number of locations. Furthermore, there may be uncertainty in the number of locations. Moreover, global topological information is rarely present. Instead, we propose a method for reasoning about location and structure within the \emph{local}, observable topology. Restricting the scope allows inference algorithms to reason about multiple topologies with varying numbers of nodes. 

\begin{figure}[!h]
  \centering
  \includegraphics[scale=0.22]{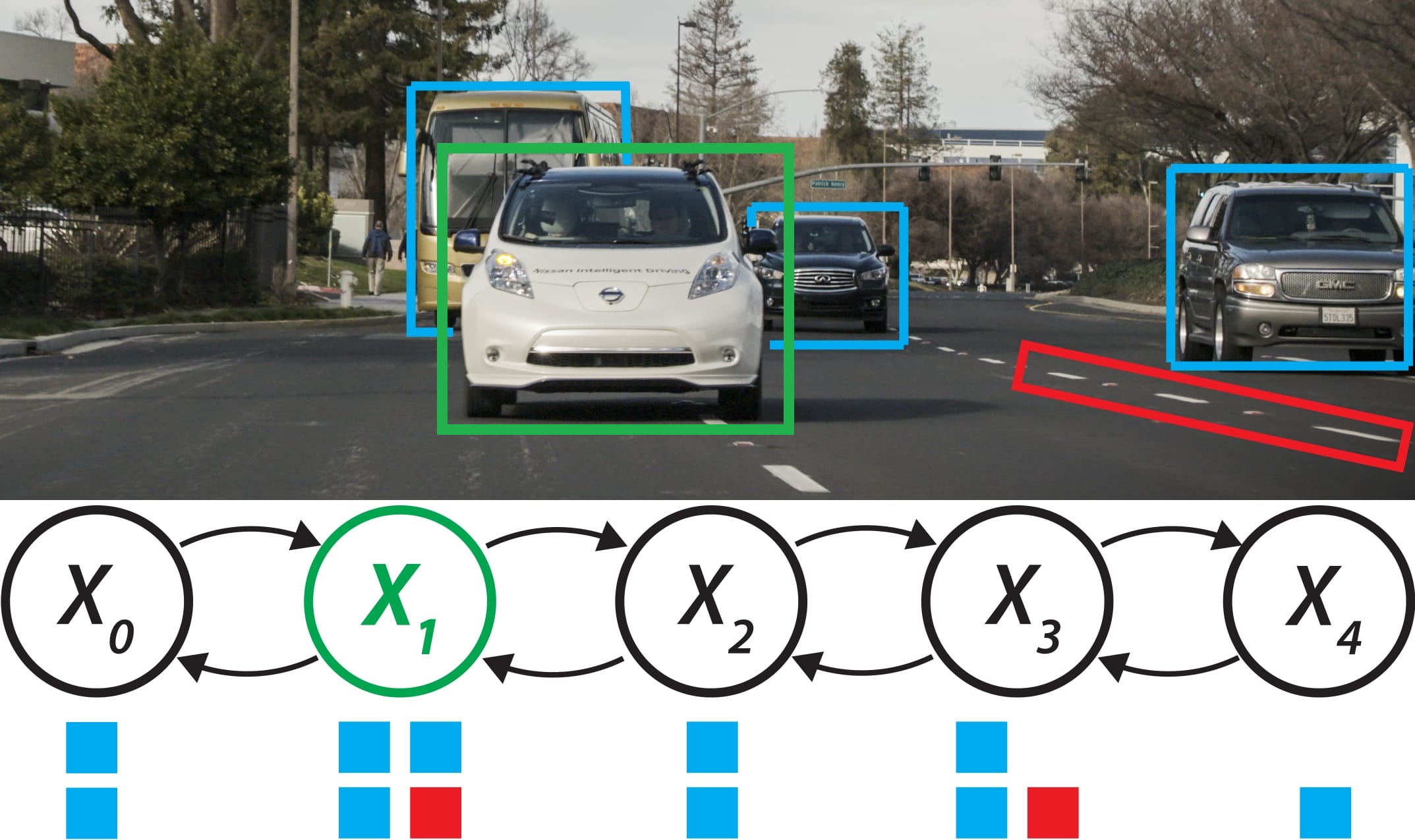}
  \caption{Agreement of observations to HMM lane-states. State $x_1$, representing being between the right and center lanes, is the only state for which all three vehicle detections (blue) and the lane line detection (red) are likely for the AV (green).}
\vspace{-7mm}
\label{fig:teaser}
\end{figure}

The discrete nature of topological location, combined with the requirement to reason about multiple possible realities simultaneously, motivates our approach. The first key idea, shown in \figref{teaser}, is to use Hidden Markov Models (HMMs) for localization by modeling expected observations and transitions at and between topological nodes. The second key idea is to use a variable set of HMMs to model a variety of possible realities. Here, HMMs model the transition dynamics and observation models of a topological map, analogous to how Kalman filters model these aspects~of a tracked object. Thus, we call this approach Variable Structure Multiple Hidden Markov Models (VSM-HMM); its function being similar to Variable Structure Multiple Models~\cite{li1996multiple}.~An important distinction is that VSMMs allow tracking of objects which follow a variety of process models, whereas the VSM-HMM approach reasons about multiple world models.

This paper presents three contributions. First, we demonstrate a method for applying HMMs to lane-level localization on an AV (\S\ref{s:hmm}). Second, we describe our VSM-HMM approach to managing a dynamic set of HMMs, each of which estimates a location within a unique local topology, allowing us to reduce dependence on high-definition (HD) maps (\S\ref{s:vsmhmm}). Third, we extend the Earth Mover's Distance (EMD) in order to handle distributions which have domains of different sizes during model belief initialization (\S\ref{s:eemd}).

Our approach is evaluated in simulation as well as on 6 real-world data sets gathered on public, multi-lane roads in Silicon Valley. Results presented in \S\ref{s:results} show that the VSM-HMM model can provide accurate topological location estimates, as well as detect disagreements between the topology specified by the map and that supported by observation.

\section{Related Work}

There are several different approaches to topological localization. The problem of global topological localization may be modeled as a Partially Observable Markov Decision Processes (POMDP). However, approaches using POMDPs, either in conjunction with geometric landmarks~\cite{zanichelli1999topological} or fingerprints from visual input~\cite{tapus2006topological}, do not scale well with the number of topological nodes. 

To reduce computational complexity, some approaches eliminate most of the reasoning about uncertainty by directly matching the robot's current view against representative feature vectors from topological locations in the map. A variety of map representations, feature extraction methods, and distance measures have been examined, including Generalized Voronoi Graphs~\cite{choset2001topological}, SIFT and SURF features~\cite{andreasson2004topological}, and Jeffrey divergence~\cite{ulrich2000appearance}, respectively. These approaches work well for environments in which nodes can be visited often, and their representative feature vectors kept up-to-date, as in~\cite{dayoub2008adaptive}. However, this is rarely possible for an AV. 

Localization algorithms specifically for AVs have typically focused on metric location, relying on high-definition (HD) maps for reliable, global data association. There are many variants, including approaches which use vanilla particle filters~\cite{schindler2013vehicle}, Rao-Blackwellized particle filters~\cite{lee2007constrained}, and Kalman Filters~\cite{schreiber2013laneloc,tao2013mapping,franke2013making}. However, these approaches require HD maps and compute both metric and topological location in a single pass, which our approach does not. 

In contrast, the proposed VSM-HMM model allows decoupling of metric and topological estimates, creating a hybrid metric-topological problem similar to~\cite{ranganathan2013light,ranganathan2013towards}, although this paper focuses only on the topological component. Similar to FastSLAM~\cite{montemerlo2002fastslam}, VSM-HMM maintains multi-modal belief over not only topological location, but also~the local topological structure. Until now, particle filters were the only viable multi-modal topological localization framework. Moreover, particle filters resampling ignores local topological structure, whereas the VSM-HMM exploits this.

Dynamic Bayes Nets~\cite{murphy2002dynamic} are common tools for localization~\cite{kaess2010bayes,biswas2014episodic}, and HMMs specifically have been used for topological localization before. In~\cite{kosecka2004vision} HMMs are used as an optional layer to process video feed in the case of a low-confidence nearest neighbor match, and in~\cite{hensel2011probabilistic} HMMs are trained to detect and classify specific railroad turnouts using sequences of signals from eddy current sensors. Multiple HMMs have been suggested for other tasks where the number of classes is high, such as genome sequencing~\cite{gough2001assignment}, and hierarchical HMMs~\cite{tishbyhierarchical} have been used for language, handwriting, and speech recognition.

\section{Lane Identification using HMMs} \label{s:hmm}

Hidden Markov Models are promising candidates for topological localization for two primary reasons. First, HMMs are well understood theoretically and support many efficient modes of inference. In our application, we use the Forward algorithm because it affords low computational cost. Second, HMMs support learning and are easily designed for specific sensor features and or topological structure. Additionally, methods such as~\cite{sivaraman2013integrated}, can be used to refine observations prior to evaluation by the HMM.

\vspace{-4mm}

\begin{figure}[!h]
  \centering
  \includegraphics[scale=0.3]{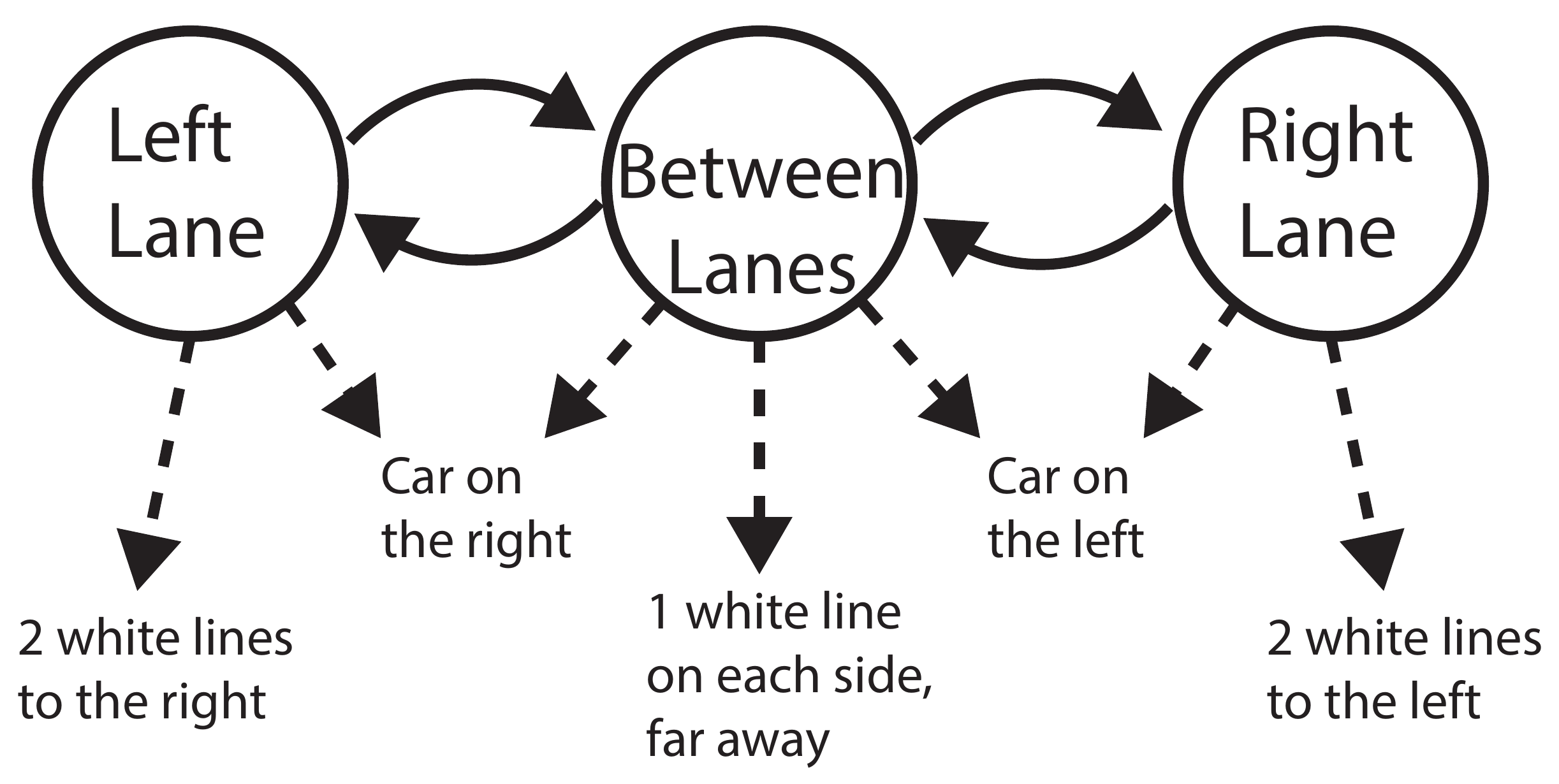}
  \caption{Simplified example lane-state HMM. Solid lines represent non-zero transition probabilities between states. Self-loops are not shown. Dashed lines represent observation or emission probabilities, some of which are exclusive for a single state. Note that the transition structure of the HMM models the topological structure of the environment. Some types of obervations have been omitted for simplicity.}
\vspace{-3mm}
\label{fig:HMM}
\end{figure}

Topological location is modeled as occupancy of a state $x_i \in X$ within an HMM. The application of this work is toward lane-level localization, and thus the states $X$ in the HMM correspond to either being in the center of a lane or being between two lanes. For example, an HMM representing a two lane road, detailed in \figref{HMM}, has three states $x_0$, $x_1$, and $x_2$. $x_0$ and $x_2$ correspond to occupying the right lane and the left lane, respectively. $x_1$ represents having some portion of the car over the lane divider. We call states like $x_1$ \emph{switching states}. In general, an HMM representing a road with $L$ lanes will have $2L - 1$ states. 

A particular strength of HMMs in topological localization is their ability to efficiently model real-world dynamics via their transition function. Since the AV can only move from one lane to an adjacent lane by moving through an immediately adjacent switching state, the transition matrix representing the transition function is sparse. We define the state transition matrix for an $n$-state HMM, ${\bf \tau^n}$, as

\vspace{-1mm}

\begin{equation}
\tau^n _{ij} = \left \{ \begin{array}{lll}
t_r \text{ if } i = j \\
t_s \text{ if } |i-j| = 1 \\
0 \text{ otherwise}, \\
\end{array} \right.
\end{equation}

\vspace{-1mm}

\noindent where $t_r$ and $t_s$ are parameters for the probability of remaining in the same state and switching to an adjacent state, respectively. This transition matrix reduces reports of physically impossible events, such as instantaneous changes across multiple lanes, which is a key advantage over other multimodal approaches such as particle filters.

\begin{figure}
  \centering
  \includegraphics[scale=0.3]{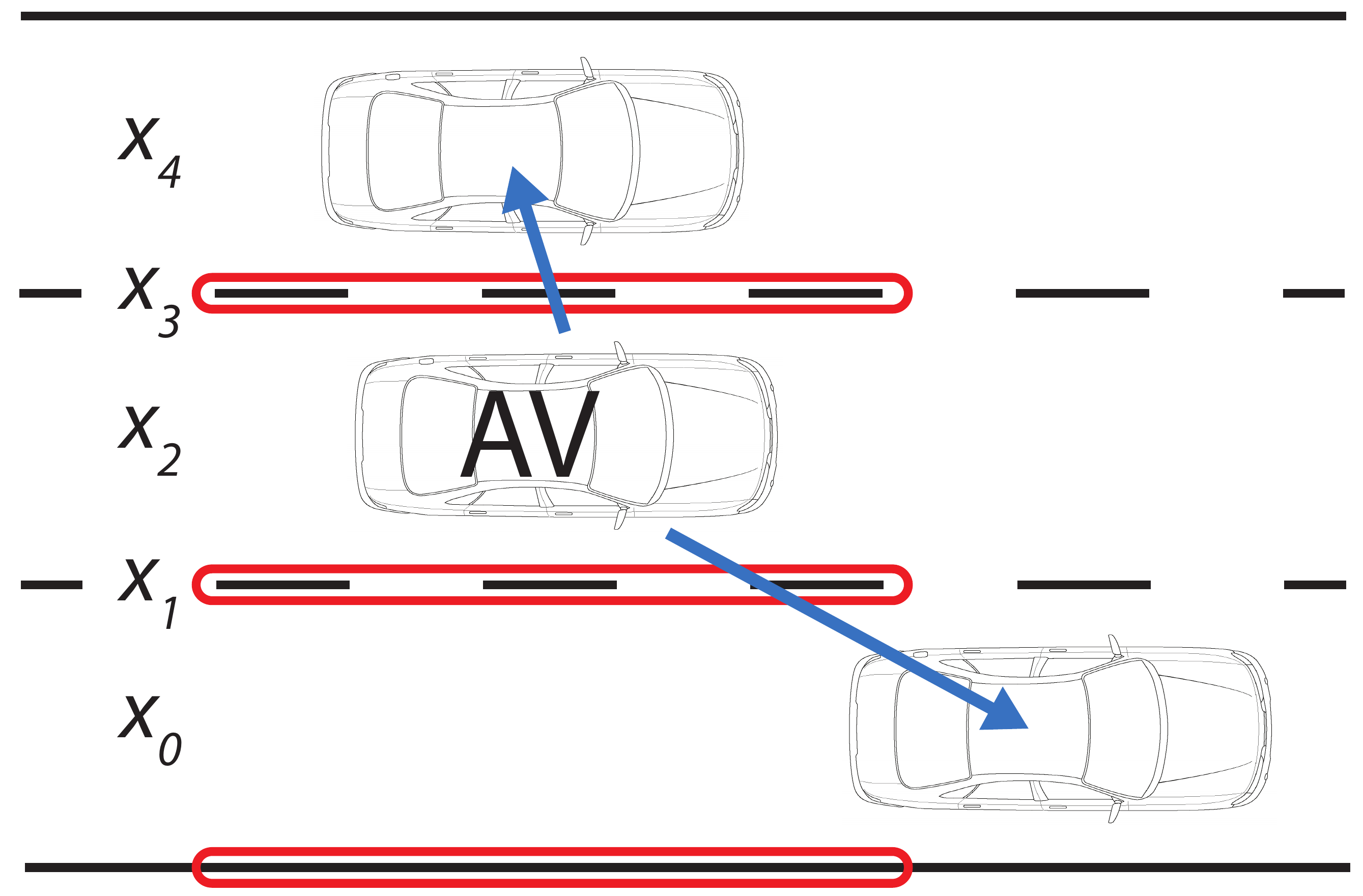}
  \caption{Diagram of lane-states and observable features. Lane-states $x_0, \dots x_4$ correspond to distinct topological regions on the road for which there are expected observations. $x_2$ is the only state which both lane line measurements (red) and vehicle detections (blue) support. Lane line measurements alone would result in equal belief in states $x_2$ and $x_4$.}
\vspace{-7mm}
\label{fig:AVobs}
\end{figure}

In addition to GPS and inertial sensors, we use lane line detections and observed relative locations of other vehicles to inform our topological estimate, shown in \figref{AVobs}. For lane lines we use a combination of learned parameters and information from a map to parametrize a Gaussian Mixture Model which describes the likelihood of observing a lane line at positions and orientations relative to the AV, given a particular lane-state.

Since observations of lane lines are unreliable due to occlusions, weather and lighting conditions, and absence of the markings themselves on many roads, we also use the relative positions of nearby tracked vehicles to support occupancy of certain lane-states. The key idea is that, although other vehicles move both globally and relatively with respect to the AV, they do so according to a specific local pattern defined by lane membership. For example, if the AV is traveling on a two-lane road and senses a vehicle immediately to its right, then there is a high probability that the AV is in the left lane, since the observed vehicle is far more likely to be traveling in the right lane than to be traveling beyond the edge of the road. We denote the observation function for state $x_i$ and sensor $q$ as $\phi(x_i, q)$.

Although single HMMs are reliable for topological localization when the map is correct and the number of states~in the HMM matches the number true lane-states, they can fail when this is not true. To deal with this, we introduce the Variable Structure Multiple Hidden Markov Model.

\section{Variable Structure Multiple HMMs} \label{s:vsmhmm}

The Variable Structure Multiple Hidden Markov Model (VSM-HMM) is similar to the variable-structure multiple-model set of Kalman Filters used in many tracking domains. However, whereas the multiple-model approaches in the tracking and control literature predict the current dynamical mode of the tracked object, the VSM-HMM approach estimates the current structure of the local topology. That is, the VSM-HMM hypothesizes about the outside world state rather than an internal process model. This is necessary when the local topological map has non-zero uncertainty.

We define a VSM-HMM as a set of all possible models $U$, an active model set $U_A$ and an inactive model set $U_N$. Every $u \in U$ is an HMM with a unique transition matrix defined by the number of lanes. $U_A \cup U_N = U$, and $U_A \cap U_N = \emptyset$.

At every time step, $U_A$ is determined using a variation of the Likely Model Set (LMS) algorithm~\cite{li2000multiple}, outlined in Algorithm~\ref{alg:lms}. For every $u \in U$, we compute a model likelihood~(line 6). Pr$(u|M)$ is the probability of model $u$ given the map $M$. If there was no uncertainty in $M$, then Pr$(u|M)$ would be 1 for the model suggested by the map, $u_M$, and 0 otherwise. In our implementation, Pr$(u|M) = \alpha 2^{-|(|u| - |u_M|)/2|}$, where $|u_M|$ is the number of states in $u_M$, and $\alpha$ is a normalizing constant. Pr$(z|u)$ is the maximum probability of observing $z$ given some state~in $u$, $\textsc{Max}_{x_i}$Pr$(z | {x_i} \in X_u)$. In \figref{AVobs}, Pr$(z|u)$ would be low for models with fewer than three lanes since $z$ contains features which are unexpected in models with fewer than three lanes.

After model likelihood is computed, up to $\kappa$ models are chosen so long as the ratio between their likelihood and the maximum likelihood of all models is above a threshold $T_{active}$ (lines 8-14). $\kappa$ is chosen based on computational constraints. Last, belief is copied from active models and initialized for inactive models (lines 15-19). Initializing belief is done using the Extended Earth Mover's Distance (\S\ref{s:eemd}).

\vspace{-1mm}

  \begin{algorithm}[ht]
  \caption{\textsc{Likely Model Set}}
  \begin{algorithmic}[1]
  \State $\textbf{Input:}$ All models $U$, active models $U_A$, observations $z$, max number of models $\kappa$, threshold $T_{active}$, map $M$
  \State $\textbf{Output:}$ Set of updated most likely models $U'_A$
  \State $U'_A \gets \emptyset$
  \State $S \gets [ \hspace{1mm} ]$
  \ForAll{$u \in U$}
  \State $\mathcal{L} \gets \text{Pr}(u|M) \times \text{Pr}(z|u)$
  \State $S \gets S.\textsc{append}(\mathcal{L}, u)$
  \EndFor
  \State $\mathcal{L}_{max} \gets Max_{\mathcal{L}}(S)$
  \ForAll{$\mathcal{L}, u \in S$}
  \If{$U'_A = \emptyset$}
  \State $U'_A \gets U'_A \cup u$
  \Else 
  \If{$|U'_A| < \kappa$ and $\frac{\mathcal{L}}{\mathcal{L}_{max}} > T_{active}$}
  \State $U'_A \gets U'_A \cup u$
  \EndIf
  \EndIf
  \EndFor
  \ForAll{$u' \in U'_A$}
  \If{$u' \in U_A$}
  \State $u' \gets \textsc{copyExistingBelief}(u')$
  \Else
  \State $u' \gets \textsc{initNewBelief}(u')$
  \EndIf
  \EndFor
  \Return $U'_A$
  \end{algorithmic}
  \label{alg:lms}
  \end{algorithm}

\vspace{-3mm}

Discrepancies between the map and reality are detected by calculating the entropy, $H$, of the posterior probability (belief) over the states in each model: 

\begin{center}
$\displaystyle H(bel(X)) = - \frac{1}{\log(|X|)} \sum_{i = 1} ^{|X|} bel(x_i) \log ( bel(x_i) )$.
\end{center} 

If the model suggested by the map has a high entropy compared to another model, the map is likely incorrect since high entropy indicates no state in the suggested topology can explain the observations. Note that the normalized equation for entropy calculates equal values for all models when no information is present. Thus, having a lack of information altogether will not cause the algorithm to flag the map as having an error. Only observations which both contradict the map's topology \emph{and} may be explained by a different topology will result in a high entropy ratio. If $|U_A| > 1$, serving a localization requests amounts to picking the most likely state from the model with the lowest entropy.

In theory, one could devise a single HMM with a dense transition matrix {\bf $\tau^*$}, which models the same problem. We can define {\bf $\tau^*$} in terms of the sub-blocks representing each distinct topological hypothesis, {\bf $\tau^1$}, {\bf $\tau^2$}, \ldots, {\bf $\tau^n$} (models in the VSM-HMM), and the transitions between them. Let each sub-block {\bf $\tau^k$} lie on the diagonal of {\bf $\tau^*$}. Further, let the off-diagonal blocks hold the probabilities of switching between sub-blocks, $t_m$. Note that $t_m$ is a notational placeholder for a range of values corresponding to the probabilities of transitioning between two specific sub-blocks (models).

\vspace{-4mm}

\begin{equation}
\tau^* _{ij} = \left \{ \begin{array}{ll}
\tau^k _{lm} \text{ if } i,j \text{ index } l,m \text{ in sub-block } k \\
t_m \text{ otherwise}. \\
\end{array} \right.
\end{equation} 

\begin{figure}
  \centering
  \includegraphics[scale=0.20]{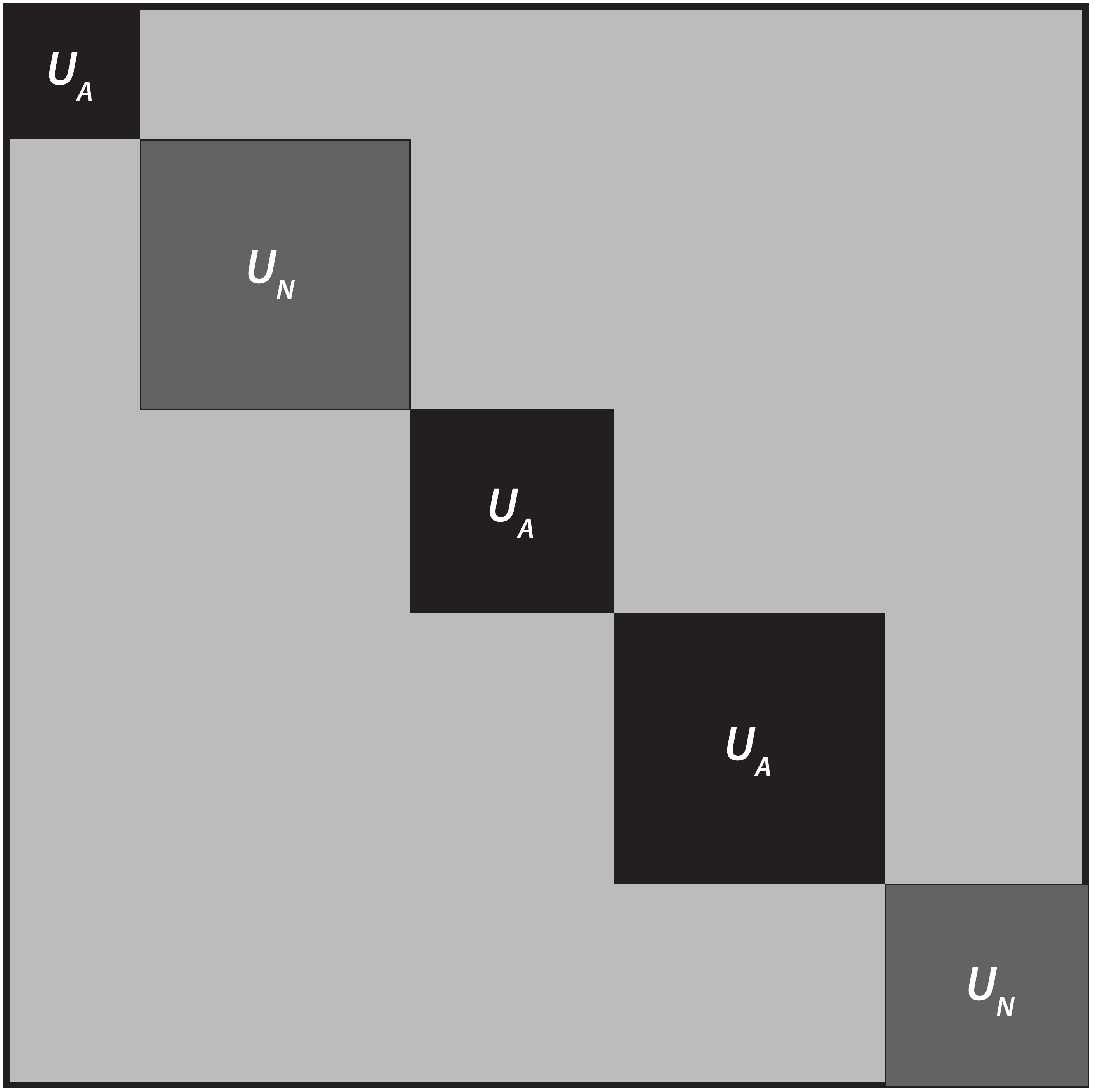}
  \caption{Transition matrix for single HMM analog to VSM-HMM. Light gray areas hold probabilities of switching models, $t_m$. Block diagonals represent all models in $U$. Note that the VSM-HMM reasons about only the active models, represented by the black sub-blocks.}
\vspace{-7mm}
\label{fig:HMManalog}
\end{figure}

In general, this results in a $p \times p$ transition matrix, where $p = \sum_{u \in U} |X_u|$. Similarly, $X^*$ and $\phi^*$ are defined by the union of all state spaces and observation functions, respectively. However, even when $U_A = U$, the VSM-HMM is more computationally efficient than its single~HMM analog since calculating Pr$(x_t|x_{t-1})$ only considers the block diagonals instead of the entire dense matrix, and calculating when to switch models depends only on the block diagonal~size. In practice, $U_A \subset U$ is much more common. In this~case, the VSM-HMM approximates the equivalent HMM by reasoning over a subset of the most likely belief points, shown in \figref{HMManalog}.

\section{Extended Earth Mover's Distance} \label{s:eemd}

Whenever a model is initialized, it needs a starting belief. Suppose an AV has a belief, $\beta$, about its current topological position in a local topology. $\beta$ is discrete and lies on the $n$-simplex, $\Delta^n$, where $n+1$ is the number of local topological states. Here, $n +1 = 2L - 1$, where $L$ is the number of lanes on the road the AV is traveling. Further, suppose the AV is nearing an intersection or merge in which the number of lanes on the road the AV will end up on is $L'$. Once on the new road the AV will initialize a new model and need a new belief, $\beta'$, about its topological location. However, if $L' \not = L$, $\beta$ and $\beta'$ will be over different numbers of states. The question is, if $L' \not = L$ how do we initialize $\beta'$, given $\beta$.

One option is to erase all previous belief and start over from uniform, $\beta' = \mathcal{U}(0,m)$. Another option is to heuristically initialize $\beta'$, such as right- or left-alignment of lane-states. Both options are computationally efficient, but do not perform optimally in many cases. A third option is to initialize $\beta'$ as the `closest' distribution in $\Delta^m$ to $\beta$, where `closeness' is defined by some statistical metric. This is preferable, but there are no metrics which satisfy the constraints of the problem since there is no isomorphism between $\Delta^n$ and $\Delta^m$, and the mapping from some belief point in $\Delta^n$ to the corresponding point $\Delta^m$ is uncertain. One example of uncertainty is a two-lane road which becomes a three-lane road across an intersection. It is unclear whether the two lanes in the first road correspond to the two rightmost, two leftmost, or some other combination of lanes in the three-lane road.

Thus, we introduce a statistical metric based on the Earth Mover's Distance (EMD)~\cite{rubner2000earth}, called the Extended Earth Mover's Distance (EEMD), which measures the \emph{expected} distance between distributions on simplices of arbitrary relative size, given the probability of all mappings between simplices. Proof of metric properties is in the Appendix. We initialize $\beta'$ such that EEMD($\beta, \beta'$) is minimized.

Before defining EEMD, we introduce some notation. Let $\mathbb{P}^n$ and $\mathbb{P}^m$ be normalized distributions on $\Delta^n$ and $\Delta^m$, respectively, and define $N = n+1$, and $M = m+1$. Without loss of generality, suppose $n > m$. Let the function ${\bf f}^{m, n}  : \Delta^m \rightarrow \Delta^n$ be defined as 

\vspace{-1mm}

\begin{equation}
f_j ^{m, n} (\mathbb{P}^m) = \left \{ \begin{array}{ll}
\mathbb{P}_j^m \text{ if } j \leq M \\
0 \text{ if } j > M. \\
\end{array} \right.
\end{equation}

\vspace{-1mm}

Thus, ${\bf f}^{m, n}$ pads $\mathbb{P}^m$ with dimensions with zero belief, making it the same size as $\mathbb{P}^n$. We denote this new distribution, now on $\Delta^n$, $\mathbb{P}^{m'}$. We can now use the original formulation of EMD to compute distance between $\mathbb{P}^{m'}$ and $\mathbb{P}^n$. However, in general, there may be uncertainty in the mapping between the two distributions. It may be that $\mathbb{P}_j ^{m'}$ and $\mathbb{P}_j ^n$ do not correspond to the same real world state. There are $N^N$ possible mappings from $\mathbb{P}^{m'}$ to $\mathbb{P}^n$. We calculate the expected distance by summing over all possibilities for $\mathbb{P}^{m'}$, and compute the EMD weighted by the probability of each mapping, Pr$(\mathbb{P}^{m_i '})$. Thus, we define the Extended Earth Mover's Distance between $\mathbb{P}^n$ and $\mathbb{P}^m$ as

\vspace{-5mm}

\begin{equation}
\text{EEMD}(\mathbb{P}^n, \mathbb{P}^m)  = \sum_{i = 1} ^{N^N} \text{Pr}(\mathbb{P}^{m_i '}) \text{EMD}(\mathbb{P}^n, \mathbb{P}^{m_i '}).
\end{equation}

\vspace{-1mm}

It is assumed that $\text{Pr}(\mathbb{P}^{m_i '})$ is known and normalized. In practice we calculate it using information from the map, and our problem has structure allowing us to ignore most of the summands since the corresponding $\text{Pr}(\mathbb{P}^{m_i '})$ term is 0. We use the EEMD as a principled guide to constructing distributions for model initialization. $\beta'$ is calculated such that EEMD($\beta, \beta'$) is minimized.

\section{Results} \label{s:results}

To test the VSM-HMM framework, we perform two experiments. The first measures localization accuracy, and the second tests the framework's ability to reason about local topological structure and detect discrepancies between the map and reality.

Localization accuracy was tested on 6 hand-annotated datasets gathered by an AV on public, multi-lane roads near Nissan Research Center in Silicon Valley. Each dataset was recorded over about a mile of stop-and-go traffic and ranged in time from 2 to 6 minutes. All road segments had between 3 and 6 lanes, corresponding to between 5 and 11 states. In these experiments no metric location information, such as GPS, was used, and topological ground truth was provided.

Because of the intermittent nature of the lane line and vehicle detections, not all timesteps possess enough observations to disambiguate lane-states. Thus, in the results presented in \tabref{results} we do not consider instances in which either no observations were recorded or the observations voted for at least half of all states, such as seeing only a single lane line immediately to the left of the vehicle. These instances are labeled ``Missing Observations". Given sequences of timesteps with little or no observations, it is possible to have multiple states tie for the same belief.  Localization is considered correct if the true state is among those with maximum belief, and incorrect otherwise. Length and observation quality are shown so as to give an idea about the difficulty of the dataset. Predictions are made at 100Hz.

\vspace{-2mm}

\begin{table}[h]
\centering
\begin{tabular}{| l | l | l | l |}
\hline
Dataset & Length (mins) & Missing Obs. & Accuracy \\
\hline
1 & 2.0 & $11 \%$ & $83 \%$ \\
2 & 3.7 & $18 \%$ & $74 \%$ \\
3 & 4.8 & $6 \%$ & $83 \%$ \\
4 & 4.4 & $9 \%$ & $95 \%$ \\
5 & 3.5 & $30 \%$ & $81 \%$ \\
6 & 5.7 & $20 \%$ & $73 \%$ \\
\hline
\end{tabular}
\vspace{-2mm}
\caption{Location estimation results}
\vspace{-3mm}
\label{tab:results}
\end{table}

\vspace{-1mm}

Testing topological structure estimation was done using simulated data, since there were too few cases in the real world data to draw concrete conclusions. To simulate false positive data, lane line and vehicle detections are generated according to the topology given by the map with probability $P_M$, and according to some other, randomly selected topology with probability $(1 - P_M)$. Further, to simulate the intermittent nature of real-world data, with probability $(1 - P_E)$ no observations are emitted. To see how our approach handles increased sensor noise, we tested different levels of variance. Given a variance $\sigma$ based on real-world data, simulated observations are generated with variance $K_{\sigma} \sigma$, where $K_{\sigma}$ is an experimental parameter.

Locations of lane lines and vehicle detections are sampled from multivariate normal distributions with means as a function of the given topology, and variances $K_{\sigma} \sigma$. The observation generation process runs independently for each lane line and vehicle detection. \tabref{simresults} displays the results of local topological structure estimation. Combined, these results demonstrate VSM-HMM as an effective framework for dealing with topological uncertainty.

\begin{table}
  \centering
  \begin{tabular}{|l|c||c|c|c|c|c|c|c|c|c|}
    \hline
    \multicolumn{2}{|c|}{$P_E$} & \multicolumn{3}{|c|}{0.9} & \multicolumn{3}{|c|}{0.7} & \multicolumn{3}{|c|}{0.5} \\
    \hline
    \multicolumn{2}{|c|}{$K_{\sigma}$} & 1 & 2 & 3 & 1 & 2 & 3 & 1 & 2 & 3 \\
    \hline \hline
    \multirow{4}{*}{$P_M$} & 0.9 & 96 & 95 & 92 & 94 & 88 & 83 & 83 & 78 & 72 \\
     & 0.8  & 87 & 85 & 80 & 84 & 79 & 72 & 63 & 60 & 54 \\
     & 0.7 & 75 & 74 & 72 & 66 & 65 & 61 & 55 & 50 & 50 \\
     & 0.6 & 66 & 63 & 63 & 62 & 63 & 59 & 51 & 48 & 49 \\
    \hline
  \end{tabular}
\vspace{-1mm}
\caption{Local topological structure estimation accuracy. Results are reported as the percent of timesteps during which the correct topological structure was estimated with highest probability (lowest entropy). $P_M$ is probability of sampling from the correct topology. $P_E$ is the probability of emitting observations. $K_{\sigma}$ is the amount by which the variance is scaled. Each entry in the table was computed from performance over 1000 timesteps.}
\vspace{-7mm}
\label{tab:simresults}
\end{table}

\section{Conclusion}

This paper presents a framework, Variable Structure Multiple Hidden Markov Models (VSM-HMM), for topological localization in the presence of topological uncertainty. We present empirical results from both simulated and real-world data on an autonomous vehicle which support VSM-HMM's effectiveness. Future work includes automating the generation and maintenance of observation and transition models through learning, as well as integrating this approach with other map representations.   

\bibliography{LuTU}

\section{Appendix}

\begin{theorem}
\emph{EEMD is a metric, having the properties of non-negativity, identity, symmetry, and the triangle inequality.} 
\end{theorem}

\noindent {\bf Non-negativity:} Since EMD is a metric, it is always positive. $\text{Pr}(\mathbb{P}^{m_i '})$ is always non-negative. Thus, their product is always non-negative, and the sum of non-negative elements is also non-negative. \\

\vspace{-2mm}

\noindent {\bf Identity:}

\noindent (EEMD$(\mathbb{P}^n, \mathbb{P}^m) = 0 \implies \mathbb{P}^n = \mathbb{P}^m)$. \\ If EEMD$(\mathbb{P}^n, \mathbb{P}^m) = 0$, then for all summands, either $\text{Pr}(\mathbb{P}^{m_i '}) = 0$ and or $\text{EMD}(\mathbb{P}^n, \mathbb{P}^{m_i '}) = 0$. Since $\text{Pr}(\mathbb{P}^{m_i '})$ is a distribution, there must be at least one $i$ such that $\text{Pr}(\mathbb{P}^{m_i '}) > 0$. If there are more than one such $i$, then EEMD$(\mathbb{P}^n, \mathbb{P}^m)$ cannot be zero, since each $\mathbb{P}^{m_i '}$ is unique and EMD is a metric, violating the assumption EEMD$(\mathbb{P}^n, \mathbb{P}^m) = 0$. If there is a single $i$ such that $\text{Pr}(\mathbb{P}^{m_i '}) > 0$, then since EMD is a metric and must be 0, EEMD$(\mathbb{P}^n, \mathbb{P}^m) = 0 \implies$ EMD$(\mathbb{P}^n, \mathbb{P}^{m_i '}) = 0 \implies \mathbb{P}^n = \mathbb{P}^m$. \\

\vspace{-2mm}

\noindent ($\mathbb{P}^n = \mathbb{P}^m \implies$ EEMD$(\mathbb{P}^n, \mathbb{P}^m) = 0)$. \\
This is clear from the definition of EEMD and $\text{Pr}(\mathbb{P}^{m_i '})$. All $\text{Pr}(\mathbb{P}^{m_i '})$ will be 0 except when $\mathbb{P}^{m_i '} = \mathbb{P}^m$. For this term, the corresponding EMD will be 0 since EMD is a metric. \\

\vspace{-2mm}

\noindent {\bf Symmetry:} The smaller dimension is always augmented, regardless of order. Thus, since $\text{Pr}(\mathbb{P}^{m_i '})$ is constant, the exact same calculation is performed for both EEMD$(\mathbb{P}^n, \mathbb{P}^m)$ and EEMD$(\mathbb{P}^m, \mathbb{P}^n)$. So, EEMD$(\mathbb{P}^n, \mathbb{P}^m)$ = EEMD$(\mathbb{P}^m, \mathbb{P}^n)$. \\

\vspace{-2mm}

\noindent {\bf Triangle Inequality:} Let $m$, $n$, and $k$ be non-negative integers, and consider the three simplices, $\Delta ^m$, $\Delta ^n$, and $\Delta ^k$. Without loss of generality, let $m < n < k$ and define $N = n+1$, $M = m+1$, and $K = k +1$. 

\vspace{-1mm}

\begin{lemma} \label{lemer}\ \\
EEMD$(\mathbb{P}^n, f^{m, n}(\mathbb{P}^m)) = $ EEMD$(f^{n, k}(\mathbb{P}^n), f^{m, k}(\mathbb{P}^m))$.
\end{lemma}

\noindent From the definition of EEMD, 

\vspace{-5mm}

\begin{equation}
\text{EEMD}(\mathbb{P}^n, f^{m, n}(\mathbb{P}^m))  = \sum_{i = 1} ^{N^N} \text{Pr}(\mathbb{P}^{m_i '}) \text{EMD}(\mathbb{P}^n, \mathbb{P}^{m_i '})
\end{equation}

\vspace{-3mm}

\noindent and

\vspace{-7mm}

\begin{equation}
\text{EEMD}(f^{n, k}(\mathbb{P}^n), f^{m, k}(\mathbb{P}^m)) \hspace{-1mm} = \hspace{-1mm} \sum_{i = 1} ^{K^K} \hspace{-1mm} \text{Pr}(\mathbb{P}^{m_i ''}) \text{EMD}(\mathbb{P}^{n'}, \mathbb{P}^{m_i ''}).
\end{equation}

We call dimensions $1$:$N$ \emph{essential} dimensions, and dimensions $(N+1)$:$K$ \emph{extra} dimensions. The sum on the RHS of equation (6) can be decomposed into two parts: one sum, with $N^N$ terms, which corresponds to all mappings in which the extra $K-N$ dimensions map only amongst themselves, and another sum, with $K^K - N^N$ terms, which corresponds to all mappings where at least one of the extra dimensions maps to one of the essential dimensions. Thus, we can rewrite the RHS of equation (6) as 

\vspace{-5mm}

\begin{equation}
\sum_{i = 1} ^{N^N} \text{Pr}(\mathbb{P}^{m_i ''}) \text{EMD}(\mathbb{P}^{n'}, \mathbb{P}^{m_i ''}) + \hspace{-3mm} \sum_{i = N^N + 1} ^{K^K} \hspace{-4mm} \text{Pr}(\mathbb{P}^{m_i ''}) \text{EMD}(\mathbb{P}^{n'}, \mathbb{P}^{m_i ''}).
\end{equation}

\noindent Since none of the extra dimensions has any meaning in the original problem and therefore cannot possibly map to any essential dimension, $\text{Pr}(\mathbb{P}^{m_i '}) = 0$ for all $i > N^N$. Thus, the second sum is 0 and equation (6) becomes

\vspace{-5mm}

\begin{equation}
\sum_{i = 1} ^{K^K} \text{Pr}(\mathbb{P}^{m_i ''}) \text{EMD}(\mathbb{P}^{n'}, \mathbb{P}^{m_i ''}) \hspace{-1mm} = \hspace{-1mm}
\sum_{i = 1} ^{N^N} \text{Pr}(\mathbb{P}^{m_i ''}) \text{EMD}(\mathbb{P}^{n'}, \mathbb{P}^{m_i ''}).
\end{equation}

\noindent Furthermore, since all extra dimensions have weight 0, their contribution to all summands in equation (8) is 0. So, equation (8) becomes 

\vspace{-5mm}

\begin{equation}
\sum_{i = 1} ^{K^K} \text{Pr}(\mathbb{P}^{m_i ''}) \text{EMD}(\mathbb{P}^{n'}, \mathbb{P}^{m_i ''}) = 
\sum_{i = 1} ^{N^N} \text{Pr}(\mathbb{P}^{m_i '}) \text{EMD}(\mathbb{P}^{n}, \mathbb{P}^{m_i '}),
\end{equation}

\noindent establishing the lemma. Now, let $\mathbb{P}^{m}$, $\mathbb{P}^{n}$, and $\mathbb{P}^{k}$ be distributions on $\Delta ^m$, $\Delta ^n$, and $\Delta ^k$, respectively. Consider  

\vspace{-2mm}

\begin{equation*}
\begin{aligned}
& \text{EEMD}(\mathbb{P}^{m}, \mathbb{P}^{k}), \text{ and } \\
& \text{EEMD}(\mathbb{P}^{m}, \mathbb{P}^{n}) + \text{EEMD}(\mathbb{P}^{n}, \mathbb{P}^{k}).
\end{aligned}
\end{equation*}

\vspace{-1mm}

\noindent By Lemma \ref{lemer}, we can rewrite these as 

\vspace{-4mm}

\begin{equation*}
\begin{aligned}
& \text{EEMD}(f^{m, k}(\mathbb{P}^{m}), \mathbb{P}^{k}), \text{ and } \\
& \text{EEMD}(f^{m, k}(\mathbb{P}^{m}), f^{n, k}(\mathbb{P}^{n})) + \text{EEMD}(f^{n, k}(\mathbb{P}^{n}), \mathbb{P}^{k}).
\end{aligned}
\end{equation*}

\vspace{-2mm}

All distributions now lie on $\Delta^k$. Since all nodes in the simplex are unit distance from all other nodes, then the distance calculated by EMD for weight moving from one node to any other node will be the magnitude of the weight. Thus, if $\mathbb{P}^{n}$ maintains belief on any nodes with different magnitude than $\mathbb{P}^{m}$ or $\mathbb{P}^{k}$, then

\vspace{-4mm}

\begin{equation*}
\begin{aligned}
& \text{EEMD}(f^{m, k}(\mathbb{P}^{m}), \mathbb{P}^{k}) < \\
& \text{EEMD}(f^{m, k}(\mathbb{P}^{m}), f^{n, k}(\mathbb{P}^{n})) + \text{EEMD}(f^{n, k}(\mathbb{P}^{n}), \mathbb{P}^{k}).
\end{aligned}
\end{equation*}

\vspace{-1mm}

\noindent Otherwise, 

\vspace{-5mm}

\begin{equation*}
\begin{aligned}
& \text{EEMD}(f^{m, k}(\mathbb{P}^{m}), \mathbb{P}^{k}) = \\
& \text{EEMD}(f^{m, k}(\mathbb{P}^{m}), f^{n, k}(\mathbb{P}^{n})) + \text{EEMD}(f^{n, k}(\mathbb{P}^{n}), \mathbb{P}^{k}).
\end{aligned}
\end{equation*}

\end{document}